%% file: acl_latex.tex
\documentclass[11pt]{article}

\usepackage[preprint]{acl}

\usepackage{times}
\usepackage{latexsym}

\usepackage[T1]{fontenc}

\usepackage[utf8]{inputenc}

\usepackage{microtype}

\usepackage{inconsolata}

\usepackage{graphicx}
\usepackage{caption}
\usepackage{subcaption}

\usepackage{booktabs}
\usepackage{threeparttable}  

\usepackage{amsmath}    
\usepackage{amssymb}    

\usepackage{colortbl}

\definecolor{AptMQL‑Bench}{HTML}{000000} 
\definecolor{nl-to-mongosh}{HTML}{000000} 
\definecolor{TEND}{HTML}{000000} 
\definecolor{DocSpider}{RGB}{0,0,0} 

\usepackage{xcolor}
\usepackage{soul}
\sethlcolor{yellow!30}
\usepackage{listings}
\usepackage{tabularx}
\usepackage{seqsplit}

\usepackage{tikz}

\usepackage{hyperref}
\usepackage{array}

\input{error_analysis/acl_visualization_macros}

\title{AptMQL‑Bench: From Text‑to‑SQL to Text‑to‑MQL via Access‑Pattern Schema Design and Data‑Preserving Migration}

\author{Hy Nguyen$^1$\thanks{Corresponding author. Work done during internship at MongoDB.}, Nabi Rezvani$^2$, \textnormal{and} Robin Vujanic$^2$ \\
\begin{tabular}{cc}
  The University of Sydney$^1$ &
  MongoDB Research$^2$ \\
  \texttt{nngu0448@uni.sydney.edu.au}$^*$ & \texttt{robin.vujanic@mongodb.com}$^\dagger$ \\
\end{tabular}
}

\begin{document}
\maketitle
\begin{abstract}
Document databases such as MongoDB are core infrastructure for modern applications, and natural-language interfaces to them---text-to-MQL---would let non-experts query complex, semi-structured data without mastering the query language.
Progress on this task depends on high-quality benchmarks, which are most practically obtained by converting an existing text-to-SQL benchmark to the document setting.
Unfortunately, existing efforts rely on heuristics for mechanical conversion: the document schema mirrors the relational foreign-key graph, and each query mirrors its source SQL. 
As a result in our experiments, these approaches fail to migrate 6 of 21 BIRD databases outright, silently drop up to 25.9\% of rows on others, and yield schemas whose ground-truth queries run over an order of magnitude slower as the data scales. 
We instead propose a conversion pipeline, driven by coding agents with human-in-the-loop verification, that designs each document schema from expected access patterns and rewrites queries to be MongoDB-native.
Applying it to BIRD, we build an access-pattern-based text-to-MQL benchmark (AptMQL‑Bench). 
It includes 21 document-oriented databases, 3,186 natural-language requests, and their associated MQL queries---whose databases are migrated from SQLite without data loss and scale efficiently. 
The strongest model, Claude Opus 4.5, achieves only 57.38\% accuracy without external knowledge evidence and 70.34\% with it.
This indicates that realistic text-to-MQL generation remains challenging.
\end{abstract}

\section{Introduction}
\begin{figure*}[!htp]
\centering
\includegraphics[width=\textwidth]{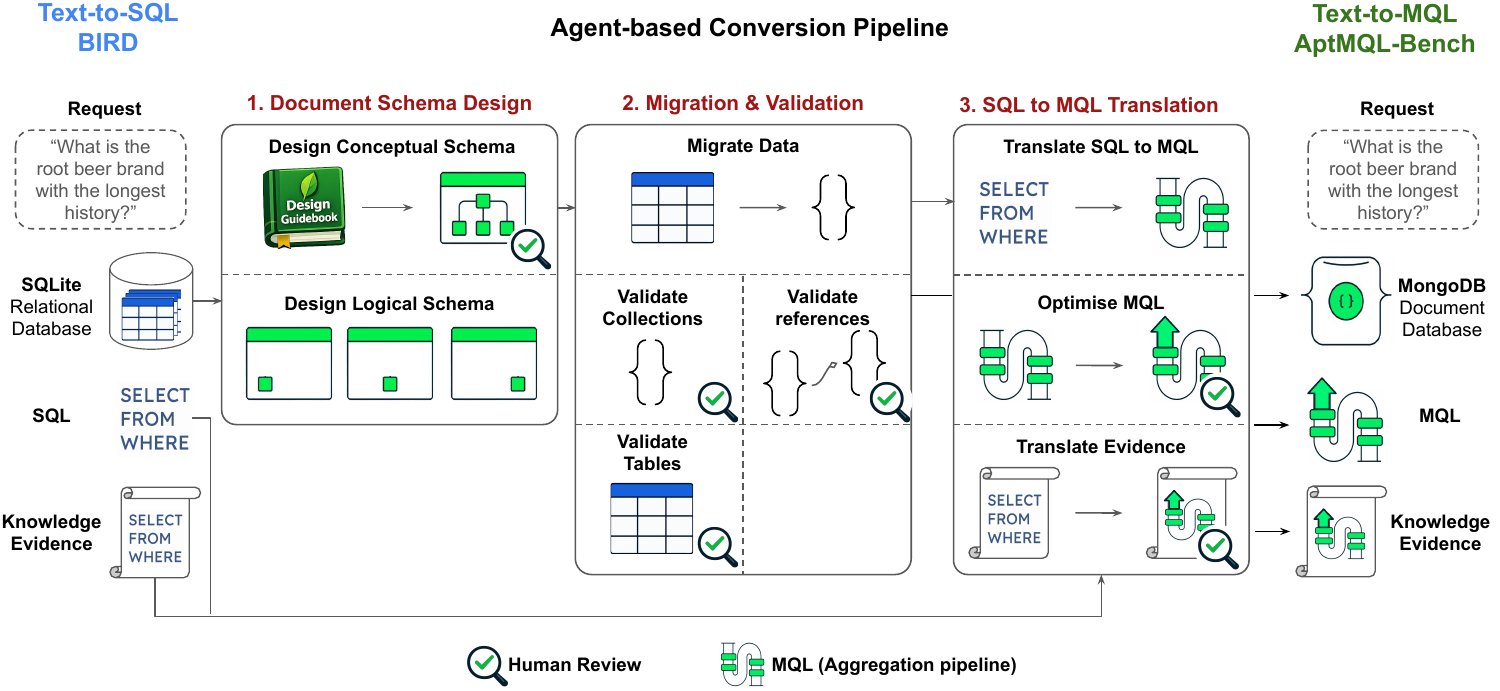}
\caption{Conversion pipeline from Text-to-SQL (left) to Text-to-MQL (right).}
\label{fig:text-to-sql-to-mql}
\end{figure*}

Text-to-MQL is the task that converts human-readable text requests into structured MQL queries (MongoDB Query Language).
A natural way to build a text-to-MQL benchmark is to convert an existing text-to-SQL benchmark into a document-database benchmark, while reusing its databases, natural-language questions, and SQL queries.
Existing conversions are largely mechanical: DocSpider~\citep{ozer_Cekinel_Toroslu_Karagoz_2025} maps each relational table to a separate collection one-to-one, while TEND-v2~\citep{lu2025tend-v2} and its multilingual extension MultiTEND~\citep{qin-etal-2025-multitend} embed tables along foreign keys.
In both cases, the MongoDB schema is derived from relational structure, and MQL is generated from gold SQL under result-equivalence supervision.\footnote{The TEND benchmark exists as two substantially different versions of the same preprint (arXiv:2502.11201). 
\emph{TEND-v2} (version 2) builds MongoDB databases by
foreign-key embedding; this is the approach its multilingual extension
\emph{MultiTEND}~\citep{qin-etal-2025-multitend} adopts. 
\emph{TEND-v3} (version 3) is a later revision that replaces this with a manually authored, MongoDB-native benchmark. 
Until now, both TEND versions remain preprints; MultiTEND, which builds on the
TEND-v2, is peer-reviewed (ACL Findings 2025).
We therefore compare against the foreign-key conversions for TEND-v2 and MultiTEND and report the differences to TEND-v3.
}

However, both mechanical conversion strategies have several limitations. 
Foreign-key embedding is especially fragile.
First, deriving document schemas from foreign-key topology alone can produce poor or even infeasible designs: high-cardinality relationships (e.g. circular reference) are difficult to embed cleanly, and some databases cannot be converted at all~\citep{lu2025tend-v2}.
In our experiments, 6 of the 21 BIRD databases~\cite{li2023can} failed to migrate outright (Table~\ref{tab:integrity}): three because a self-referential foreign key triggered infinite recursion, and three produced individual documents so large that they did not fit in MongoDB's 16 MB document limit.
Second, the conversion can silently lose data by dropping rows whose foreign keys reference no surviving parent, with row loss reaching 25.9\% in the worst case (Table~\ref{tab:integrity}). 
Third, both conversions are inefficient: one-to-one conversion joins tables with unindexed cross-collection \texttt{\$lookup}s, while foreign-key embedding makes queries \texttt{\$unwind} entire nested arrays. 
These methods result in computationally inefficient MQL queries, with inefficiencies that worsen as data scales, as shown in section~\ref{sec:schema-comparison}.

We propose a conversion pipeline that treats schema design as a first-class step and then revises the schema during the migration of the database, as shown in Figure~\ref{fig:text-to-sql-to-mql}. 
The pipeline is driven by coding agents (e.g. Claude Code\footnote{\href{https://claude.com/product/claude-code}{https://claude.com/product/claude-code}}) and uses human-in-the-loop verification at each validation stage to prevent hallucinations and data loss. 
We instantiate our pipeline on MongoDB, a widely used document database, and its query language MQL, with the following stages.
First, it designs document schemas from expected access patterns. 
Second, it migrates the databases and checks that no data is lost. 
Third, it translates SQL queries into MQL and then optimises the MQL to remove SQL-shaped artefacts.
We apply this pipeline to BIRD to build AptMQL‑Bench, a benchmark comprising 21 document-oriented databases, 3,186 natural-language requests, and their associated MQL queries.
Its databases are migrated without data loss, and its ground-truth MQL queries remain efficient even as the databases grow large.

We also evaluate proprietary and open-source LLMs. 
The strongest proprietary model, Claude Opus 4.5, achieves 70.34\% soft execution accuracy, while its open-source counterpart, GPT-OSS-120B, reaches only 62.46\%. 
These results show that text-to-MQL generation remains challenging. 
We release the benchmark, prompts, and evaluation scripts to support future work.

\section{Related Work}
\textbf{Text-to-SQL.}
Translating natural language questions into executable SQL has been a long-standing focus of semantic parsing research. 
There are many large-scale benchmarks such as WikiSQL~\cite{zhongSeq2SQL2017}, Spider~\cite{yu-etal-2018-spider}, Spider 2.0~\cite{ICLR2025_46c10f6c}, BIRD~\cite{li2023can}, and BIRD-Interact~\cite{huo2026birdinteractreimaginingtexttosqlevaluation}, which have progressively raised the bar on schema complexity, query realism, and real-world context.
Recent work leverages LLMs through few-shot prompting, domain-specific fine-tuning, and test-time scaling strategies such as agentic pipelines that inspect schemas and iteratively refine queries~\cite{shkapenyuk2025automatic,wang2025agentar,ICLR2025_974ff7b5,zhai-etal-2025-optimizing}, achieving strong results. 

\textbf{Relational-to-NoSQL conversion.}
Converting relational databases into document or other NoSQL models has been studied extensively in the data-management community, along two broad lines. 
The first is structure-driven: rule-based transformations map tables, keys, and relationships (association, aggregation, inheritance) directly to nested or referenced documents from the relational schema alone~\citep{data4040148, kuderu_relational_2016, fouad2019model}. 
The second is workload-driven, arguing that good NoSQL schemas should be derived from the expected query workload rather than relational structure.
NoSE~\citep{mior2017nose} casts this as cost-based schema optimisation over a conceptual model and query set, and subsequent work generalises the idea to document and other aggregate-oriented stores via query-pattern models~\citep{chen2022workload}. 
Our pipeline adopts a workload-driven view for benchmark construction: each document schema is designed based on expected access patterns rather than foreign-key topology, in contrast to the structure-driven embedding used by TEND-v2~\citep{lu2025tend-v2} and MultiTEND~\citep{qin-etal-2025-multitend}.

\textbf{Text-to-NoSQL Benchmarks.}
Natural-language querying of NoSQL databases has received far less attention than text-to-SQL. 
Several benchmarks are built by converting a relational source:
DocSpider~\citep{ozer_Cekinel_Toroslu_Karagoz_2025} maps each Spider table to a separate collection one-to-one, and TEND-v2~\citep{lu2025tend-v2}, with its multilingual extension MultiTEND~\citep{qin-etal-2025-multitend} embeds tables along foreign keys.
Moreover, they generate MQL from gold SQL under result-equivalence supervision, and their databases inherit relational structure rather than document-native design.
Others are built manually: a later version TEND-v3~\citep{lu2025tend-v3} has expert authors of MongoDB-native databases, and SM3-Text-to-Query~\cite{sivasubramaniam2024sm3} hand-designs a multi-model benchmark limited to the medical domain.
Our work occupies the middle ground: we use coding agents under human-in-the-loop verification to convert an existing text-to-SQL benchmark---preserving its databases, questions, and SQL---through access-pattern-driven schema design and migration without data loss across domains.

\section{The Conversion Pipeline}
\label{sec:pipeline}
The pipeline is driven by a coding agent (e.g. Claude Code) with lightweight human review at designated quality gates, and proceeds in three stages: (1) MongoDB schema design, (2) database migration and validation, and (3) SQL-to-MQL translation, as illustrated in Figure~\ref{fig:text-to-sql-to-mql}.
Rather than performing the conversion as a single end-to-end task, we decompose each stage into smaller subtasks, each executed by the coding agent.

\subsection{Schema Design}
\label{sec:schema-design}

Following the classical organisation of database design into conceptual, logical, and physical levels~\citep{1855347}, we split schema design into a \emph{conceptual} design that fixes the collections and their relationships and a \emph{logical} design that fixes the field-level structure of each collection.
An example of conceptual and logical schema designs can be found in Appendix~\ref{app:schema-design}.

\textbf{Design guidebook.}
Both conceptual and logical designs are grounded in a MongoDB design guidebook that we curate from three official MongoDB University courses\footnote{\href{https://learn.mongodb.com/}{https://learn.mongodb.com/}}.
The guidebook is central to our approach for three reasons. 
First, it lays out a step-by-step procedure for designing a document schema from a relational one, distilling the practice recommended by MongoDB experts. 
Second, it establishes the \textit{Golden Rule}---\textit{data that is accessed together is stored together}---as the principle every design decision must follow, as illustrated in Figure~\ref{fig:golden-rule}. 
Third, it describes and exemplifies the design patterns MongoDB advocates (e.g. computed, extended reference, subset, polymorphic) and the anti-patterns to avoid (unbounded arrays, bloated documents), equipping the model with a standard set of modelling decisions.

\begin{figure}[t]
\centering
\begin{lstlisting}[basicstyle=\ttfamily\footnotesize, frame=single,
                   columns=fullflexible, breaklines=true]
{
  "_id": "10012", 
  "brandName": "Fitz's",
  "rating": {          // computed pattern
    "average": 4.3, "reviewCount": 157
  },
  "featuredReviews": [ // subset pattern (top 3)
    { 
      "reviewId": "101811",
      "reviewerName": "Kenneth Walton",
      "starRating": 5, 
      "reviewText": "Fantastic!" 
    }
  ]
}
\end{lstlisting}
\caption{A document in a \texttt{brands} collection from \texttt{beer\_factory}, a DB related to a beer retail shop. 
On the website, a brand page shows the brand together with its overall rating and a few sample reviews, so these are stored together: the rating is precomputed (computed pattern) and the top reviews embedded (subset pattern). Because the full review set is unbounded, it is kept in a separate \texttt{reviews} collection (reference).}
\label{fig:golden-rule}
\end{figure}

\textbf{Conceptual Schema Design}
Given the source SQLite database and its column descriptions, a coding agent session is prompted to follow the guidebook and produce a design document with four parts. 
The first is an analysis of the source relational schema, the application domain, and a data profile for each entity (sizes, cardinalities, and value distributions). 
The second states the assumed access patterns and operation rates that the design must support. 
The third records the modelling decisions: for each relationship, the embedding-versus-referencing choice and its rationale, together with the guidebook patterns applied and anti-patterns avoided. 
The fourth lists the resulting collections and indexes, and the field-level transformations from SQLite to MongoDB. 
The design is then checked in two ways. 
First, a reviewer inspects and confirms that each modelling choice is reasonable. 
Second, the reviewer also verifies that the design loses no data on the conceptual level: every source table must remain retrievable from the proposed design document.

\textbf{Logical Schema Design}
\label{sec:logical-design}
The conceptual schema specifies \textit{what} collections exist and how they relate, but not the precise field-level structure of each collection; the second sub-task fills in this detail. For each collection, we invoke a separate agent session with the SQLite database, the design guidebook, the conceptual schema design, and the target collection name. 
The agent locates every part of the conceptual design that references the target collection, queries the SQLite database for source columns and types, and defines all fields that should appear in the collection. 
The output is a CSV file in which each row corresponds to a single field, with four columns: the source SQLite column(s), the target MongoDB field, the MongoDB data type, and transformation notes.

\subsection{Migration and Validation}
\label{sec:migration}

Migration is not a row-by-row copy but the application of the design from Section~\ref{sec:schema-design}, which restructures the source data.
For example, several SQLite tables may collapse into one collection, or a single row may be duplicated across embedded copies and extended references, and values are retyped. 
This restructuring is where errors creep in. 
This is why migration and validation must go hand in hand to make these errors detectable and overcome LLM shortcomings.

\textbf{Data Migration}.
\label{sssec:percollection-migration}
For each collection, we run an agent session with the SQLite database, the conceptual schema, and the logical schema of the target collection.
From these, the agent produces a \emph{migration} Python script that applies the logical schema and bulk-inserts the data into MongoDB.

\textbf{MongoDB Collection Validation}
At this step, the coding agent is prompted to produce a \emph{validation} script comprising a list of validation mappings derived from the logical design.
Each validation mapping pairs an SQL query over the source SQLite table with an MQL query over the migrated MongoDB collection; the two reference the same columns and fields and must return identical results (Appendix~\ref{app:validation-mappings}).
Because the mappings are anchored on the logical design, together they confirm that every field the design specifies was migrated as intended.
This check matters because, when a collection has many fields, the agent often migrates and validates most of them but leaves a few behind, dropped through type mismatches, unexpected values, or overlooked columns.
The validation script makes such omissions detectable: any field whose mapping fails to validate is flagged, and the reviewer then inspects the failure and directs the agent to resolve it (e.g., fixing the migration code, then migrating and validating again).
This process iterates until every field in the logical design validates.

\textbf{MongoDB Cross-collection Validation}.
We also validate the integrity of cross-collection references introduced by the document design.
For example, in the code snippet at Figure~\ref{fig:golden-rule}, all items in \texttt{featuredReviews} must exist in the \texttt{reviews} collection. 
For each collection, the coding agent identifies every such reference and emits validation mappings that pair source- and target-side aggregation pipelines whose outputs must agree under containment.
A reviewer then inspects the generated mappings and runs them to confirm reference integrity.
If any cross-collection reference fails to hold, the reviewer works with the agent to resolve the mismatch, iterating until every reference holds.

\textbf{SQLite Table Validation.}
The collection validation of the previous step checks each collection against the source it was designed from, so it cannot catch tables or columns that never entered the design in the first place.
While finalising the schema, the agent may fail to enumerate every column of a wide table and drop some, or judge a table's data obsolete and exclude it altogether.
We therefore add a source-anchored pass that walks the SQLite side directly, using a predefined script.
For every SQLite column, the script infers its most likely MongoDB destination and flags any source value with no match as probable data loss. 
A flag may be a genuine loss or merely an intentional transformation, such as a retyped or renamed value.
The agent therefore inspects each flag, querying both the SQLite table and the MongoDB collection to compare the actual values, and classifies it as either an intentional transformation or a genuine loss.
The reviewer then examines only the genuine losses and works with the agent to update the logical schema of the target collection, rerun the migration, and rerun the validations, repeating until every SQLite table is fully accounted for.

\subsection{Translation from SQL to MQL}
\label{sec:translation}

\textbf{Initial SQL-to-MQL Translation}
The third stage translates each natural language--SQL pair into a corresponding natural language--MQL pair.
Regarding the NL request, we retain the original.
For each SQL query, we invoke a coding agent session with the SQL query, the migration scripts, and a predefined query-result comparison script, and prompt it to produce a corresponding translation to MQL.
The agent iterates on the pipeline until exact result equivalence with the original SQL is verified; cases that cannot be translated without unnatural constructs are flagged for manual review.

\textbf{MQL Optimisation}
Translations that pass result-equivalence validation are correct but often retain a relational shape (SQL legacy). 
For example, performing redundant \texttt{\$lookup} stages that look up already-embedded documents (e.g. the \texttt{rating} field in Figure~\ref{fig:golden-rule}).
We therefore add an optimisation pass: the coding agent inspects each translated MQL against the document model, identifies these SQL legacy patterns, and rewrites the pipeline to leverage embedded fields and precomputed aggregates directly. 
A reviewer then collaborates with the agent to manually check each optimised MQL to confirm it is semantically equivalent to the original translation.
In total, 1,444 MQLs (45.32\%) were optimised in this way.

\textbf{External Knowledge Translation}
Some text-to-SQL benchmark pairs each natural language question with an associated external-knowledge field---free-form text that supplies the domain-specific information, such as the definition of a domain-specific metric or technical term.
For each evidence, we invoke a coding agent session with the natural language request, the BIRD evidence, the SQL query, the ground-truth MQL, and the database schema, prompted to rewrite the evidence so that field references use MongoDB \texttt{collection.field} paths drawn from the MQL and operational hints describe the corresponding MongoDB-idiomatic behaviour in natural language. 
A reviewer inspects each translated evidence string to confirm that field references and operational hints align with ground-truth MQL.

\section{AptMQL‑Bench}

We build AptMQL‑Bench by applying the conversion pipeline to 21 databases from BIRD~\citep{li2023can}, ten and eleven databases from BIRD's train and development splits, respectively. 
We summarize the statistics of AptMQL‑Bench and other text-to-MQL datasets in Table~\ref{tab:dataset_stats}\footnote{We report the two TEND releases separately: TEND-v2~\citep{lu2025tend-v2} converts databases by foreign-key embedding but publishes only its migration code, not the resulting natural-language requests or MQL queries, whereas TEND-v3~\citep{lu2025tend-v3} is the later, expert-authored release.}.

\paragraph{Databases}
AptMQL‑Bench contains 21 document databases spanning the diverse domains of BIRD.
Its databases are the most structurally complex among comparable benchmarks, averaging 7.1 collections per database---more than the manually authored NL-to-MongoSH~\citep{mongodb_eai_nl_to_mongosh}, SM3-Text-to-Query~\citep{sivasubramaniam2024sm3}, and TEND-v3~\citep{lu2025tend-v3}, and than the flat one-to-one conversion of DocSpider~\citep{ozer_Cekinel_Toroslu_Karagoz_2025}---and roughly 218k documents per database. 
Moreover, the benchmarks differ in construction method: the databases in NL-to-MongoSH, TEND-v3, and the single-domain SM3-Text-to-Query are authored manually (H), while DocSpider and TEND-v2 convert relational sources mechanically by program (P) and thus inherit relational rather than document-native structure. 
AptMQL‑Bench is converted by coding agents with human review (A+H) and migrates all 21 databases without data loss (Table~\ref{tab:integrity}).

\paragraph{Natural-language requests and MQL queries.}
AptMQL‑Bench pairs its databases with 3,186 natural-language requests, reused verbatim from BIRD, so that user intent is expressed independently of the query language. 
Averaging 82 characters with a 3-gram diversity of 0.68, they are the most lexically varied requests among all benchmarks in Table~\ref{tab:dataset_stats}. 
AptMQL‑Bench's MQLs are also comparatively complex: from single-intent requests of 82 characters on average, they reach 4.0 aggregation stages at a nesting depth of 4.8, exceeding NL-to-MongoSH (2.81 stages, 4.01 depth) and DocSpider (3.3, 3.79).

\paragraph{Human Effort}
Because human review is integral to the pipeline, we give an approximate accounting of the time it required, estimated by the reviewer rather than instrumented (Appendix~\ref{app:reviewer}). 
Schema design and validation—conceptual design together with the three validation passes—took on the order of 27 minutes per database on average (roughly 13–42 min), dominated by conceptual design (around 12 min); across the eleven development databases. 
This is on the order of five person-hours. 
On the translation of SQL-to-MQL, the initial translation is verified automatically under result-equivalence and needs little manual time; manual effort concentrates in the optimisation pass (30–60 seconds per optimised MQL) and evidence rewriting (15–30 seconds per evidence).

\begin{table*}[!ht]
\centering
\resizebox{\textwidth}{!}{
\begin{tabular}{l|cccc|cccc|cccc}
\toprule
 \multicolumn{1}{c}{\textbf{}} & \multicolumn{4}{c}{\textbf{Databases}} & \multicolumn{4}{c}{\textbf{NL Requests}} & \multicolumn{4}{c}{\textbf{MQLs}} \\
\cmidrule(lr){2-5} \cmidrule(lr){6-9} \cmidrule(lr){10-13}
\textbf{Benchmark} & \# DB & Design by & \# Coll./DB & \# R.Docs/DB & \# NL & Length* & Diversity & Annot. & \# MQL & Length* & Depth* & Annot. \\
\midrule
NL-to-MongoSH        & 8 & H & 2.8 & 45k  & 766 & 89 & 0.52 & L & 766 & 2.81 & 4.01 & L \\
DocSpider            & 159 & P & 5.1 & 10k  & 4,663 & 70 & 0.53 & L+H & 4,663 & 3.3 & 3.79 & H \\
TEND-v2$^\dagger$              & 154 & P  & 2.2 & 1k & 17k & &  & L+H & 17k  &  & & L+H \\
SM3-Text-to-Query    & 1 & H & 5 & 1.8k & 8k & 82 & 0.19  & P+H & 8k  & 4.54 & 3.18  & P+H \\
TEND-v3              & 11 & H & 2.9 & 24k  & 1210 & 395 & 0.40 & L+H & 1210 & 7.28 & 8.14 & L+H \\
\midrule
AptMQL‑Bench  & 21 & A+H  & 7.1 & 218k & 3,186 & 82 & 0.68 & H & 3,186 & 4.00 & 4.80 & A+H \\
\bottomrule
\end{tabular}
}
\caption{
Statistics of AptMQL‑Bench and related text-to-MQL benchmarks. The "Design by" and "Annot." columns indicate how each benchmark's databases and its NL/MQL annotations were produced---by a \textbf{P}rogram using template synthesis, an \textbf{L}LM, a coding \textbf{A}gent, or a \textbf{H}uman (combinations joined by ``+''). \#Coll./DB and \#R.Docs/DB are the average numbers of collections and root documents per database. 
For NL requests, \emph{Length} is the average character count and
\emph{Diversity} the ratio of distinct to total 3-grams~\citep{li-etal-2016-diversity} under
GPT-2 tokenisation~\citep{radford2019language}; for MQLs, \emph{Length} is the average number of aggregation stages and \emph{Depth} the average nesting depth.
$^\dagger$TEND-v2 does not release NL requests and MQLs.
}
\label{tab:dataset_stats}
\end{table*}

\section{Evaluation Metrics}
\label{sec:evaluation-metrics}

Besides \emph{execution success rate (Exec-Rate)} and \emph{execution accuracy (Strict-EX)}, two commonly used metrics, we propose \emph{soft execution accuracy (Soft-EX)} as our main metric for semantic correctness, relaxing the requirement to allow the prediction to carry extra fields.
For example, rather than projecting only the \texttt{brandName} field (ground truth), a prediction may return the entire \texttt{brands} documents without any additional operators—a natural advantage of document databases in answering such queries.
Writing $G' \sqsubseteq P'$ for ``every column of $G'$ has a value-equivalent column in $P'$'', the instance-level score is
\begin{equation}
\text{Soft-EX}(P, G) = 1[G' \sqsubseteq P']
\end{equation}
and the final metric is its mean over all test instances.
Crucially, the relaxation applies only to fields, not to rows: $P'$ and $G'$ must contain the same number of rows and each ground-truth column must match a prediction column value-for-value, so a query that simply returns an entire collection does not satisfy Soft-EX unless the ground-truth answer already spans it.
We further verify that the relaxation does not over-credit: on the strongest model, every instance that Soft-EX accepts but Strict-EX rejects returns the correct answer together with additional document fields, never extra rows or disagreeing values (Appendix~\ref{app:soft-ex}).

\section{Benchmarking}
We benchmark eight LLMs on AptMQL‑Bench---five open-source models (Llama-3.1-70B, GPT-OSS-20B, GPT-OSS-120B, Qwen-3.5-35B, and Gemma-4-31B) and three proprietary Claude models (Haiku, Sonnet, and Opus 4.5), each evaluated with and without BIRD's external knowledge.
Table~\ref{tab:results} reports execution success rate (Exec-Rate), execution accuracy (Strict-EX), and soft execution accuracy (Soft-EX).

The benchmark is challenging for every model.
The strongest model, Claude-Opus-4-5, reaches only 57.38\% Soft-EX without external knowledge and 70.34\% with it.
Accuracy tracks model capability: within the Claude family, Soft-EX rises from Haiku (47.71/64.72) to Sonnet (52.01/67.86) to Opus (57.38/70.34), and the strongest open-source model, GPT-OSS-120B (44.51/62.46), trails Opus by around 8 points with knowledge.
External knowledge helps substantially and consistently, adding between 12.15 and 18.51 points (+12.96 for Opus), indicating that models cannot supply BIRD's domain-specific hints on their own.

Regarding execution success rate, it is near-saturated for all capable models (around 98.5\% for the Claude models), confirming that invalid MQL is rare and that the remaining difficulty is semantic rather than syntactic.
Strict-EX is uniformly and substantially lower than Soft-EX---for Opus, 38.45/50.44 versus 57.38/70.34, a gap of roughly 19--20 points---because it penalises any prediction that returns extra fields.
As we show in Appendix~\ref{app:soft-ex}, this gap consists entirely of predictions that return the correct answer together with additional document fields, never extra rows or wrong values, so Strict-EX is a conservative lower bound and the true accuracy lies between the two metrics, closer to Soft-EX.
The two metrics largely agree on model ordering, though the top of the strict ranking differs: Opus leads on Soft-EX, whereas Gemma-4-31B attains the highest Strict-EX with knowledge (52.39), reflecting that stronger models more often return richer---but still correct---documents that Strict-EX rejects.

\begin{table*}[t]
\centering
\definecolor{secgray}{gray}{0.92}
\definecolor{humangreen}{RGB}{200,240,200}
\resizebox{\textwidth}{!}{
\begin{tabular}{l cc cc cc}
\toprule
\textbf{Models} & \multicolumn{2}{c}{\textbf{Exec-Rate}} & \multicolumn{2}{c}{\textbf{Strict-EX}} & \multicolumn{2}{c}{\textbf{Soft-EX}} \\
\cmidrule(lr){2-3}\cmidrule(lr){4-5}\cmidrule(lr){6-7}
 & w/o knowledge & w/ knowledge & w/o knowledge & w/ knowledge & w/o knowledge & w/ knowledge \\
\midrule
Llama-3.1-70B & 80.57 & 82.80 {\scriptsize\color{gray}(+2.23)} & 15.44 & 24.48 {\scriptsize\color{gray}(+9.04)} & 24.92 & 37.07 {\scriptsize\color{gray}(+12.15)} \\
GPT-OSS-20B & 81.36 & 85.88 {\scriptsize\color{gray}(+4.52)} & 22.25 & 36.94 {\scriptsize\color{gray}(+14.69)} & 32.49 & 51.00 {\scriptsize\color{gray}(+18.51)} \\
Qwen-3.5-35B & 91.56 & 93.72 {\scriptsize\color{gray}(+2.16)} & 20.37 & 31.83 {\scriptsize\color{gray}(+11.46)} & 39.17 & 56.06 {\scriptsize\color{gray}(+16.89)} \\
Gemma-4-31B & \textbf{99.00} & \textbf{99.09} {\scriptsize\color{gray}(+0.09)} & 38.20 & \textbf{52.39} {\scriptsize\color{gray}(+14.19)} & 46.74 & 61.96 {\scriptsize\color{gray}(+15.22)} \\
GPT-OSS-120B & 96.80 & 97.83 {\scriptsize\color{gray}(+1.03)} & 29.76 & 44.19 {\scriptsize\color{gray}(+14.43)} & 44.51 & 62.46 {\scriptsize\color{gray}(+17.95)} \\
\midrule
Claude-Haiku-4-5 & 98.34 & 98.87 {\scriptsize\color{gray}(+0.53)} & 23.79 & 34.75 {\scriptsize\color{gray}(+10.96)} & 47.71 & 64.72 {\scriptsize\color{gray}(+17.01)} \\
Claude-Sonnet-4-5 & 97.93 & 98.43 {\scriptsize\color{gray}(+0.50)} & 29.44 & 41.87 {\scriptsize\color{gray}(+12.43)} & 52.01 & 67.86 {\scriptsize\color{gray}(+15.85)} \\
Claude-Opus-4-5 & 98.27 & 98.27 {\scriptsize\color{gray}(+0.00)} & \textbf{38.45} & 50.44 {\scriptsize\color{gray}(+11.99)} & \textbf{57.38} & \textbf{70.34} {\scriptsize\color{gray}(+12.96)} \\
\bottomrule
\end{tabular}
}
\caption{Performance of open-source and proprietary LLMs.}
\label{tab:results}
\end{table*}

\section{Schema Design Matters: One-to-One, FK-Embedding, and Access-Pattern Modelling}
\label{sec:schema-comparison}

\begin{figure*}[t!]
\centering
\includegraphics[width=0.95\textwidth]{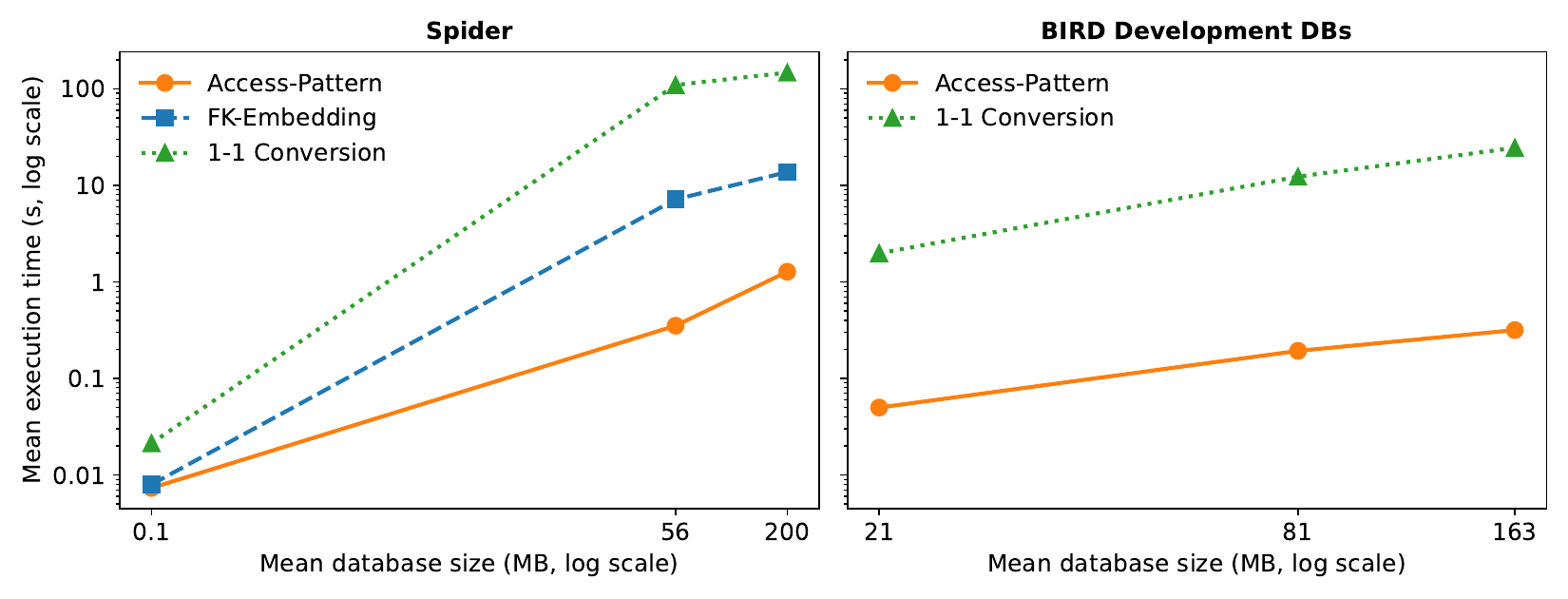}
\caption{Execution time of ground-truth MQLs versus database size (log--log), averaged across databases.
}
\label{fig:exec-scale}
\end{figure*}

When a text-to-SQL benchmark is converted as a text-to-MQL benchmark, the choice of document schema affects two properties: whether the source data is preserved and whether the ground-truth queries remain efficient as the database grows.  
We compare three conversion strategies: (1) \emph{One-to-One}, the flat table-to-collection mapping of
DocSpider~\citep{ozer_Cekinel_Toroslu_Karagoz_2025}; (2) \emph{FK-Embedding}, the foreign-key-driven embedding
of TEND-v2~\citep{lu2025tend-v2}; and (3) our \emph{Access-Pattern} document modelling.
The comparison is shown in Table~\ref{tab:integrity} for data integrity and Figure~\ref{fig:exec-scale} for query performance under scale.

\paragraph{Data integrity.}
A conversion that drops records is not faithful: the resulting questions can reference data absent from the target store.
Table~\ref{tab:integrity} reports row-level loss for the three strategies.
One-to-One is lossless by construction---each table becomes its own collection, so every row and column is copied verbatim---but the resulting databases merely mirror the relational schema and are not document-native, at a query cost we quantify below.
FK-Embedding is far less safe\footnote{For the eleven BIRD databases, we obtain FK-Embedding results from a faithful re-implementation of TEND-v2's database migration algorithm, validated to reproduce its three released conversions (\texttt{bike\_1}, \texttt{car\_1}, \texttt{pets\_1}) exactly. This supports the row-level data-loss comparison in Table~\ref{tab:integrity}; it does not supply ground-truth MQL queries for the FK-Embedding schema, required by the execution-time study (Appendix~\ref{app:scaling}).}.
It fails outright on six databases (N): \texttt{codebase\_community}, \texttt{beer\_factory}, and \texttt{food\_inspection\_2} have self-referential foreign keys, a cyclic case TEND-v2 cannot handle and discards~\citep{lu2025tend-v2}; \texttt{financial}, \texttt{european\_football\_2}, and \texttt{work\_cycles} collapse into single root documents, causing errors.
On eight more, it migrates only partially, silently dropping rows whose foreign key resolves to no surviving parent---up to 25.9\% on \texttt{toxicology}. 
These failures follow from deriving the schema from foreign-key topology alone, with no regard for cardinality or access patterns. 
Access-Pattern, which decides embedding versus referencing per relationship, migrates databases without data loss.

\paragraph{Ground-truth MQL execution under scale.}
Schema design also determines how efficiently ground-truth queries run.
This matters for the benchmark itself: even a correct MQL can become expensive on a poorly designed schema.
Timing this requires, for each strategy, the converted databases paired with ground-truth MQLs.

Both baselines---One-to-One (DocSpider) and FK-Embedding (TEND-v2)---are developed on Spider, but they differ in what they release.
DocSpider publishes all of its converted databases and queries, as well as the prompt it uses to convert SQL to MQL, making its conversion method fully reproducible.
TEND-v2, by contrast, releases ground-truth MQLs for only three sample databases (\texttt{bike\_1}, \texttt{car\_1}, and \texttt{pets\_1}) and does not release its conversion prompt, so its method cannot be reproduced beyond those three databases.
On Spider we therefore compare all three strategies on \texttt{bike\_1}, \texttt{car\_1}, and \texttt{pets\_1}.
To test at a larger and more realistic scale, we extend the study to the eleven BIRD databases, comparing Access-Pattern against One-to-One only: we reproduce One-to-One on BIRD by following DocSpider's released conversion prompt once the converted databases are built, whereas TEND-v2's method cannot be reproduced.
For every database, we apply our own pipeline to obtain the Access-Pattern MongoDB database, natural-language requests, and MQLs.
For each strategy we run that database's ground-truth MQLs (around 30 per Spider database, 50 per BIRD database) while scaling it---preserving schema, keys, and referential integrity---across three sizes: for Spider, its original size and roughly 50 and 200~MB; for BIRD, scale factors of 1, 5, and 10 (mean sizes of roughly 20, 80, and 160~MB).
Figure~\ref{fig:exec-scale} reports execution time against database size on log--log axes, averaged (geometric mean) across databases; the scaling procedure and per-database results are detailed in Appendix~\ref{app:scaling}.

One-to-One conversion is slowest throughout, because each table lives in its own collection and every query spanning a relationship performs an unindexed cross-collection \texttt{\$lookup}.
FK-Embedding is faster where relationships can be nested---it co-locates related rows so that queries \texttt{\$unwind} arrays instead of joining---but its advantage is uneven: on many-to-many schemas it cannot embed and collapses back onto cross-collection joins, tracking One-to-One (most visibly on \texttt{pets\_1}; see Appendix~\ref{app:scaling}).
Access-Pattern is fastest on every database and scales most gently, because it embeds what is accessed together, references what is not, and indexes the fields queries filter on.
On Spider at 200~MB it runs about an order of magnitude ($\sim$$11\times$) faster than FK-Embedding and roughly two orders of magnitude ($\sim$$10^2\times$) faster than One-to-One.
On BIRD, its speedup over One-to-One grows with data size, from around 40x on the original databases to 77x once they reach around 160 MB.

\section{Conclusion}
In this paper, we propose a semi-automated method for translating existing text-to-SQL benchmarks into text-to-MQL benchmarks with the assistance of coding agents. 
The pipeline consists of three main stages: (1) document schema design , (2) database migration and validation, and (3) SQL-to-MQL translation.
Applying the pipeline to BIRD, we built AptMQL‑Bench: 21 document databases, 3,186 natural-language requests and their associated MQL queries, migrated with no data loss.

Our experiments show that schema design is decisive. 
An access-pattern design migrates every database faithfully and keeps ground-truth queries efficient as the data scales.
Benchmarking eight open-source and proprietary LLMs, we find the task remains challenging: the strongest model, Claude-Opus-4-5, reaches only 57.38\% Soft Execution Accuracy without external knowledge and 70.34\% with it. 
We release the benchmark, conversion pipeline, and evaluation scripts, and hope they support future research on text-to-MQL generation.

\section*{Limitations}
We acknowledge several limitations of this work.

\paragraph{Schema design rests on assumed access patterns.}
Our pipeline designs each document schema from expected access patterns, following the principle that data accessed together should be stored together. 
To obtain these patterns, we prompt the coding agent to first infer a plausible real-world application that the database would serve and then derive the access patterns and operation rates that application would generate; a human reviewer then checks this reasoning. 
The resulting schema is therefore only as representative as the assumed workload.
Grounding schema design in practical query workloads, rather than assumed ones, is a natural direction for future work.

\paragraph{Estimation of Human Effort}
Our reporting of reviewer effort (Appendix~\ref{app:reviewer}) is an estimate rather than an instrumented measurement, and is intended only to indicate the rough scale of the human involvement.

\paragraph{Scope to a single document store} 
Our experiments focus on a single document database, MongoDB, and its query language, MQL; we do not empirically validate that the pipeline generalises to other document stores, which we leave to future work.

\section*{Ethics Statement}
This work is built upon the BIRD benchmark~\cite{li2023can}, a publicly available text-to-SQL dataset.
We use it in accordance with its intended use and original licensing terms, and no additional personal or sensitive data was collected during this study.
We will release the benchmark, translation pipeline, and evaluation scripts under CC BY-SA 4.0 to support reproducibility and encourage future research on text-to-MQL generation. 
We foresee no direct ethical risks arising from the release of these artifacts.


\bibliography{custom}

\appendix

\newcommand{\lossless}{\textcolor{teal}{\checkmark}}   
\newcommand{\notmig}{\textcolor{red}{$\times$}}

\begin{table*}[t]
\centering
\small
\setlength{\tabcolsep}{6pt}
\begin{tabular}{l r r c c c}
\toprule
\textbf{Database} & \textbf{\#Tables} & \textbf{\#Records}
 & \textcolor{DocSpider}{\textbf{One-to-One}}
 & \textcolor{TEND}{\textbf{FK-Embedding}}
 & \textcolor{AptMQL‑Bench}{\textbf{Access-Pattern}} \\
\midrule
\multicolumn{6}{@{}l}{\textit{\textbf{BIRD training databases}}}\\
beer\_factory            &  7 & 14.0k & \lossless & \notmig   & \lossless\\
cs\_semester             &  5 & 197   & \lossless & \lossless & \lossless\\
food\_inspection\_2      &  5 & 701k  & \lossless & \notmig   & \lossless\\
hockey                   & 22 & 96.1k & \lossless & 9.60\%       & \lossless\\
mondial\_geo             & 34 & 21.1k & \lossless & 1.01\%     & \lossless\\
professional\_basketball &  9 & 44.8k & \lossless & \lossless & \lossless\\
public\_review\_platform & 15 & 991k  & \lossless & \lossless & \lossless\\
restaurant               &  3 & 19.3k & \lossless & \lossless & \lossless\\
shooting                 &  3 & 812   & \lossless & \lossless & \lossless\\
works\_cycles            & 65 & 719k  & \lossless & \notmig   & \lossless\\
\midrule
\multicolumn{6}{@{}l}{\textit{\textbf{BIRD development databases}}}\\
financial              &  8 & 1.08M & \lossless & \notmig         & \lossless\\
european\_football\_2  &  7 & 223k  & \lossless & \notmig         & \lossless\\
codebase\_community    &  8 & 741k  & \lossless & \notmig         & \lossless\\
toxicology             &  4 & 49.8k & \lossless & \textbf{25.87\%} & \lossless\\
thrombosis\_prediction &  3 & 16.0k & \lossless & 4.61\%           & \lossless\\
california\_schools    &  3 & 29.9k & \lossless & 0.70\%           & \lossless\\
card\_games            &  6 & 803k  & \lossless & 0.01\%           & \lossless\\
student\_club          &  8 & 42.5k & \lossless & 0.007\%           & \lossless\\
superhero              & 10 & 10.6k & \lossless & \lossless       & \lossless\\
debit\_card\_special.  &  5 & 423k  & \lossless & \lossless       & \lossless\\
formula\_1             & 13 & 514k  & \lossless & \lossless       & \lossless\\
\midrule
\multicolumn{6}{@{}l}{\textit{\textbf{Spider databases}}}\\
bike\_1 & 4 & 22.2k & \lossless & \lossless & \lossless\\
car\_1  & 6 & 890   & \lossless & 0.56\%     & \lossless\\
pets\_1 & 3 & 40    & \lossless & \lossless & \lossless\\
\midrule
\textit{Migrated (lossless)} & & & 24 (24) & 18 (10) & 24 (24)\\
\bottomrule
\end{tabular}
\caption{Row-level data loss under three conversions: One-to-One
(\textcolor{DocSpider}{\textbf{DocSpider}}), FK-Embedding (\textcolor{TEND}{\textbf{TEND-v2}}),
and Access-Pattern (\textcolor{AptMQL‑Bench}{\textbf{ours}}).
\lossless{} marks a lossless migration, \notmig{} marks a database that could not be migrated at
all, and a number presents the percentage of source rows lost. 
The bottom three rows are databases used in the execution-time study in Figure~\ref{fig:exec-scale} and are derived from the Spider benchmark~\cite{yu-etal-2018-spider}; the rest are the BIRD training and development databases of AptMQL‑Bench.}
\label{tab:integrity}
\end{table*}

\section{Generality beyond MongoDB.}
We instantiate our approach on MongoDB, a widely used document database,
designing document schemas from expected access patterns, migrating the
source data without loss, and translating gold SQL into aggregation
pipelines. 
However, the overall pipeline is, in principle, applicable to other document-oriented NoSQL databases with a similar document and aggregation model (e.g., Amazon DocumentDB and Azure Cosmos DB); an empirical study of cross-store transfer is left to future work.

\section{Error Analysis}
Table~\ref{tab:error_categories} summarises our manual analysis of 100 incorrect Opus predictions with external MongoDB knowledge. 
A substantial fraction of mismatches are not clear semantic failures: 37 cases are acceptable predictions, where the predicted MQL gives a defensible answer but differs from the gold query because of arbitrary tie-breaking, benchmark-specific conventions, SQL-style row preservation, or document-level versus row-level output. 
Among genuine errors, the largest category is wrong calculation (33 cases), typically caused by changes in aggregation level, distinctness, denominator, boundary conditions, or arithmetic formula. 
We also observed 12 cases of misunderstanding of knowledge and evidence.
The remaining errors come from misunderstanding database content (8 cases), wrong schema linking (6 cases), and syntax errors (4 cases). This indicates that most failures are semantic or evidence-grounding issues rather than invalid MQL syntax.

Summary of error categories can be found in Table~\ref{tab:error_categories} while examples can be found in Table~\ref{tab:acceptable_prediction_dev_1289},
~\ref{tab:misunderstanding_database_content_dev_0871}, ~\ref{tab:misunderstanding_knowledge_evidence_dev_1052}, ~\ref{tab:wrong_schema_linking_dev_1419}, and ~\ref{tab:wrong_calculation_dev_0452}.

\begin{table*}[t]
\centering
\small
\resizebox{\textwidth}{!}{
\begin{tabular}{lrp{0.62\linewidth}}
\toprule
\textbf{Category} & \textbf{Count} & \textbf{Typical Cause} \\
\midrule
Syntax Error & 4 & Fail to execute the MQLs due to syntax errors.\\
\addlinespace
Acceptable Prediction & 37 &
Semantically reasonable prediction. \\
& & Gold uses arbitrary tie-breaks, benchmark-specific conventions, or SQL-style row preservation. Returns the answer at an incompatible row or document level.\\
\addlinespace
Misunderstanding Knowledge Evidence & 12 &
Fails to understand or apply the provided MongoDB evidence. \\
& & Often ignores explicit collection paths, field mappings, value definitions, or aggregation instructions. \\
\addlinespace
Misunderstanding Database Content & 8 &
Misinterprets field meanings, categorical values, or stored data representations. \\
& & Often involves domain-specific conditions or dataset-specific value meanings not resolved by the evidence. \\
\addlinespace
Wrong Schema Linking & 6 &
Uses the wrong collection, embedded path, join source, or field. \\
& & Often selects a related but non-equivalent entity or document path. \\
\addlinespace
Wrong Calculation & 33 &
Changes the intended aggregation level, distinctness, or denominator. \\
& & Also includes wrong boundary conditions, row preservation, or arithmetic formulas. \\
\bottomrule
\end{tabular}
}
\caption{Error categories over 100 manually analysed Opus predictions with external knowledge.}
\label{tab:error_categories}
\end{table*}

\section{Reviewer Expertise and Effort}
\label{app:reviewer}
The human-in-the-loop verification throughout our pipeline (Section~\ref{sec:pipeline}) was carried out by an internal reviewer with substantial coding and database experience across both sides of the conversion. 
On the document side, the reviewer has over two years of hands-on experience with MongoDB and holds document-schema design certificates from
MongoDB University. 
On the relational side, the reviewer has over five years of experience with relational databases, including MySQL and SQLite, which underpins the source-side validation and the SQL-to-MQL equivalence checks. 
This background spans the relational source and the document target, enabling the reviewer to assess schema design decisions, migration integrity, and query equivalence at each quality gate.

Because human review is integral to our pipeline, we give an approximate accounting of the time it required. 
These figures were estimated by the reviewer rather than instrumented, so they should be read as indicative of scale rather than exact measurements. 
Table~\ref{tab:reviewer-effort} reports the schema-side effort—conceptual design together with the three validation passes—over the eleven development databases. 
It averages roughly 27 minutes per database (about 13–42 minutes) and is dominated by conceptual design ($\sim$12 min); the validation passes are comparatively light ($\sim$3, $\sim$6, and $\sim$5 min for collection, table, and cross-collection validation). 
Across the eleven databases this amounts to on the order of five person-hours. 
Translation of SQL-to-MQL review is measured per sample rather than per database and is therefore not included in the table: the initial SQL-to-MQL translation requires negligible manual time, since the agent iterates against a result-equivalence check automatically, whereas the manual effort concentrates in two places—the MQL optimisation pass (30-60 seconds per optimised query) and evidence rewriting (15–30 seconds per sample). 
The optimisation pass is the single largest component of reviewer effort, since it is applied to every optimised query. 
Overall, the per-database schema review is modest, while the per-sample translation review scales with benchmark size.

\begin{table}[t]
\centering
\small
\setlength{\tabcolsep}{2pt}
\begin{tabular}{lrrrrr}
\toprule
Database & Conc. & Coll. & Tbl. & X-Coll. & Total \\
\midrule
california\_schools        & 15 &  3 &  5 &  2 & 25 \\
card\_games                & 13 &  4 & 22 &  3 & 42 \\
codebase\_community        & 16 &  5 &  9 &  7 & 37 \\
debit\_card\_specializing  & 14 &  3 &  3 &  4 & 24 \\
european\_football\_2      & 11 &  3 & 10 &  5 & 29 \\
financial                  & 30 &  3 &  4 &  5 & 42 \\
formula\_1                 & 12 &  6 &  6 & 12 & 36 \\
student\_club              & 10 &  3 &  2 &  7 & 22 \\
superhero                  &  5 &  2 &  3 &  3 & 13 \\
thrombosis\_prediction     &  5 &  2 &  3 &  3 & 13 \\
toxicology                 &  6 &  3 &  3 &  1 & 13 \\
\midrule
Mean                       & 12.5 & 3.4 & 6.4 & 4.7 & 26.9 \\
\bottomrule
\end{tabular}
\caption{Estimated reviewer effort (minutes per database) for schema design and validation over the eleven
development databases: \textbf{Conc.}~= conceptual design, \textbf{Coll.}~= collection validation,
\textbf{Tbl.}~= table (source-anchored) validation, \textbf{X-Coll.}~= cross-collection validation.}
\label{tab:reviewer-effort}
\end{table}

\section{Example Conceptual Schema Design}
\label{app:schema-design}

Figure~\ref{fig:schema-design-example} and Table~\ref{tab:brands-mapping} show an abridged conceptual and logical design for the \texttt{beer\_factory} database.
Table~\ref{tab:conceptual-loss-check} shows the data loss checking at the conceptual design stage.

\begin{figure*}[t]
\centering
\begin{lstlisting}[basicstyle=\ttfamily\scriptsize, frame=single,
                   columns=fullflexible, breaklines=true]
# MongoDB Schema Design - Beer Factory (abridged)
Source: beer_factory.sqlite (7 tables, 13,716 rows) -> MongoDB

## 1. Source schema, domain, and per-entity analysis
Tables: customers (554), rootbeerbrand (24), rootbeer (6,430),
transaction (6,312), rootbeerreview (713), location (3), geolocation (3).
Relationships: brand->rootbeer (1:N), brand->review (1:N),
customer->transaction/review (1:N), location<->geolocation (1:1),
rootbeer->transaction (1:1). Composite PK: rootbeerreview(CustomerID, BrandID).
Notes: booleans stored as TEXT; many nulls in social fields;
RootBeerID is unique per transaction (effectively 1:1).

## 2. Access patterns and operation rates
Read-heavy workload (~90% reads). Browse brand catalog ~1000/s;
brand detail + reviews ~200/s; customer history ~50/s.
Writes: process transaction ~20/s; submit review ~2/s.

## 3. Modelling decisions (embed vs. reference, with patterns)
- location <-> geolocation (1:1): EMBED geolocation into location.
- rootbeer -> transaction (1:1): EMBED the item into transaction.
- brand -> review (1:N, unbounded): REFERENCE (separate collection),
  plus SUBSET (top-3 featuredReviews), COMPUTED (rating avg/count),
  EXTENDED REFERENCE (brand/reviewer names on the review).
- customer -> transaction (1:N): REFERENCE + EXTENDED REFERENCE.
- Anti-patterns avoided: unbounded arrays, bloated documents.

## 4. Resulting collections, indexes, and transformations
Collections: brands, customers, locations, reviews, transactions, inventory.
Eliminated: geolocation (embedded), rootbeer (embedded).
Indexes: brands.brandName; reviews{brandId, reviewDate};
customers.contact.email (unique).
Transformations: integer IDs -> string _id; composite PK -> "cid_bid";
TEXT 'TRUE'/'FALSE' -> boolean; nulls preserved.
\end{lstlisting}
\caption{Abridged conceptual schema design document for \texttt{beer\_factory},
as produced by the schema-design stage (Section~\ref{sec:schema-design}). One
representative item is shown per part.}
\label{fig:schema-design-example}
\end{figure*}

\begin{table}[t]
\centering
\footnotesize
\setlength{\tabcolsep}{3pt}
\begin{tabularx}{\columnwidth}{@{}llXc@{}}
\toprule
SQLite table & Disposition & Target collection(s) & Ret. \\
\midrule
customers      & Standalone & \texttt{customers}                        & \checkmark \\
rootbeerbrand  & Standalone & \texttt{brands}                           & \checkmark \\
rootbeerreview & Standalone & \texttt{reviews}, \texttt{brands}         & \checkmark \\
transaction    & Standalone & \texttt{transactions}                     & \checkmark \\
location       & Standalone & \texttt{locations}                        & \checkmark \\
geolocation    & Embedded   & \texttt{locations}                        & \checkmark \\
rootbeer       & Referenced & \texttt{transactions}, \texttt{inventory} & \checkmark \\
\bottomrule
\end{tabularx}
\caption{Conceptual-level data-loss check for \texttt{beer\_factory}: every SQLite
table maps to a destination in the proposed design (\textsc{standalone},
\textsc{embedded}, or \textsc{referenced}), so all source data remains retrievable.
Column-level checking happens during migration validation
(Section~\ref{sec:migration}).}
\label{tab:conceptual-loss-check}
\end{table}

\newcommand{\fld}[1]{{\ttfamily\footnotesize\seqsplit{#1}}}

\begin{table*}[t]
\centering
\small
\setlength{\tabcolsep}{5pt}
\renewcommand{\arraystretch}{1.2}
\begin{tabularx}{\textwidth}{@{}
  >{\raggedright\arraybackslash}p{3.0cm}
  >{\raggedright\arraybackslash}p{3.6cm}
  l X@{}}
\toprule
Source (SQLite) & Target (MongoDB) & Type & Transformation \\
\midrule
\fld{rootbeerbrand.BrandID}   & \fld{brands.\_id}      & String & \texttt{INTEGER}$\rightarrow$\texttt{String} \\
\fld{rootbeerbrand.BrandName} & \fld{brands.brandName} & String & Direct mapping \\
\addlinespace
\multicolumn{2}{@{}l}{\fld{brands.rating}} & Document & Computed pattern (from \texttt{rootbeerreview}) \\
\fld{rootbeerreview.StarRating} & \fld{brands.rating.average}     & Number & \texttt{AVG(StarRating)} grouped by \texttt{BrandID}; null if no reviews \\
\fld{rootbeerreview.*}          & \fld{brands.rating.reviewCount} & Number & \texttt{COUNT(*)} grouped by \texttt{BrandID}; 0 if no reviews \\
\addlinespace
\multicolumn{2}{@{}l}{\fld{brands.featuredReviews}} & Array & Subset pattern: top 3 by \texttt{StarRating}, then \texttt{ReviewDate} \\
\fld{rootbeerreview.CustomerID, BrandID} & \fld{brands.featuredReviews.reviewId}    & String & Composite \texttt{CustomerID\_BrandID}; references \texttt{reviews.\_id} \\
\fld{rootbeerreview.CustomerID}          & \fld{brands.featuredReviews.customerId}  & String & \texttt{INTEGER}$\rightarrow$\texttt{String} \\
\fld{customers.First, customers.Last}    & \fld{brands.featuredReviews.reviewerName} & String & Concatenated \texttt{First}\,+\,\texttt{Last}, joined via \texttt{CustomerID} \\
\fld{rootbeerreview.StarRating}          & \fld{brands.featuredReviews.starRating}  & Number & Direct mapping \\
\fld{rootbeerreview.Review}              & \fld{brands.featuredReviews.reviewText}  & String & Direct mapping; nullable \\
\bottomrule
\end{tabularx}
\caption{SQLite-to-MongoDB logical design for the \texttt{brands} collection,
covering the fields shown in Figure~\ref{fig:golden-rule}.}
\label{tab:brands-mapping}
\end{table*}

\section{Validation Mappings}
\label{app:validation-mappings}
Each collection is validated by a list of \emph{validation mappings}. 
A mapping has a name and description, a \texttt{sqlite} side (a SQL query over the source, plus optional value transforms), and a \texttt{mongodb} side (the target collection and an aggregation pipeline). 
Both sides are executed and their result sets compared for equality; the mapping passes only if they agree. We keep each mapping small and single-purpose so a reviewer can confirm at a glance what it checks. A collection's mappings together cover its core fields and, where applicable, its extended-reference, computed, and subset fields.

Figures~\ref{fig:validation-example} and~\ref{fig:validation-core} show two mappings for the \texttt{brands} collection: a core-field check and a computed-field check. 
During migration, each brand's average rating and review count are precomputed from the source \texttt{rootbeerreview} table and stored on the brand document. 
The mapping recomputes the same aggregate directly from the source and compares it against the stored value, so a stale or miscomputed rating is caught. 
This illustrates why result-equivalence validation is necessary: the migration transforms the data rather than copying it, and only re-deriving the value from the source confirms the transformation is correct.

\begin{figure}[t]
\centering
\begin{lstlisting}[language=Python, basicstyle=\ttfamily\scriptsize,
                   frame=single, breaklines=true, columns=fullflexible, showstringspaces=false]
{
  'name': 'Computed Rating Validation',
  'description': 'brands.rating precomputed from rootbeerreview',

  'sqlite': {
    'query': """
      SELECT CAST(BrandID AS TEXT) AS BrandID,
             AVG(StarRating) AS avg_rating,
             COUNT(*)        AS review_count
      FROM rootbeerreview
      GROUP BY BrandID
      ORDER BY BrandID
    """,
    'transforms': {}
  },

  'mongodb': {
    'collection': 'brands',
    'aggregation_pipeline': [
      {'$match': {'rating.reviewCount': {'$gt': 0}}},
      {'$project': {
        'BrandID': '$_id',
        'avg_rating': '$rating.average',
        'review_count': '$rating.reviewCount',
        '_id': 0}},
      {'$sort': {'BrandID': 1}}
    ]
  }
}
\end{lstlisting}
\caption{A validation mapping for the \texttt{brands} collection. The SQL side
recomputes each brand's average rating and review count from the source
\texttt{rootbeerreview} table; the MongoDB side reads the precomputed
\texttt{rating} subdocument. The mapping passes only if the stored aggregate
matches the value recomputed from source, validating the computed pattern.}
\label{fig:validation-example}
\end{figure}

\begin{figure}[t]
\centering
\begin{lstlisting}[language=Python, basicstyle=\ttfamily\scriptsize,
                   frame=single, breaklines=true, columns=fullflexible, showstringspaces=false]
{
  'name': 'Core Fields Validation',
  'description': 'core brand fields (subset shown)',

  'sqlite': {
    'query': """
      SELECT CAST(BrandID AS TEXT) AS BrandID,
             BrandName,
             FirstBrewedYear,
             Caffeinated
      FROM rootbeerbrand
      ORDER BY BrandID
    """,
    'transforms': {'Caffeinated': 'boolean'}   # 'TRUE'/'FALSE' -> bool
  },

  'mongodb': {
    'collection': 'brands',
    'aggregation_pipeline': [
      {'$project': {
        'BrandID': '$_id',
        'BrandName': '$brandName',
        'FirstBrewedYear': '$firstBrewedYear',
        'Caffeinated': '$properties.caffeinated',   # nested field
        '_id': 0}},
      {'$sort': {'BrandID': 1}}
    ]
  }
}
\end{lstlisting}
\caption{A core-field validation mapping for the \texttt{brands} collection (a
subset of fields is shown). The SQL side reads the raw source columns; the MongoDB
side projects the migrated fields, including one nested under a subdocument
(\texttt{properties.caffeinated}). A \texttt{boolean} transform converts the SQLite
text value (\texttt{'TRUE'}/\texttt{'FALSE'}) so the two sides compare equal,
checking both the type conversion and the nesting.}
\label{fig:validation-core}
\end{figure}

\section{Soft-EX vs.\ Strict-EX: Analysis of the Disagreement Set}
\label{app:soft-ex}

Soft-EX (Section~\ref{sec:evaluation-metrics}) credits a prediction when every ground-truth field is reproduced, allowing the prediction to carry extra fields.
A natural worry is that this relaxation over-credits: could a prediction be scored correct simply by returning more than the ground truth?
To test this, we pair Soft-EX with Strict-EX that requires bidirectional containment---$G' \sqsubseteq P'$ \emph{and} $P' \sqsubseteq G'$, i.e.\ the two result sets are equal up to field renaming and row order---and we inspect every case where the two metrics disagree.

\paragraph{Setup.}
We take the strongest model, Claude-Opus-4-5 with external knowledge, and collect all instances that Soft-EX scores correct but Strict-EX scores incorrect.
This disagreement set contains \textbf{634} instances (Opus has 2{,}241 soft-correct and 1{,}607 Strict-correct predictions).

\paragraph{The gap is entirely ``correct answer plus extra fields.''}
Table~\ref{tab:soft-ex-stats} summarises the 634 instances.
In every case the predicted result has the same number of rows as the ground truth and reproduces all of its fields---Soft-EX already enforces exact row-count equality and value-for-value agreement on the matched fields---and differs \emph{only} by carrying additional fields.
None of the 634 differ by row count, and none contain a disagreeing value.
The extra content is typically small (a median of one extra field, such as the aggregate a query ranked by), with a long tail of predictions that return whole documents.
Strict-EX rejects all of these because the prediction is richer than the gold projection, which is precisely the natural behaviour of a document store: returning an entire document, or the value used for filtering alongside the requested field, is idiomatic and correct.
Strict-EX therefore acts as a conservative lower bound, and the true accuracy lies between Strict-EX and Soft-EX, closer to Soft-EX.

\begin{table*}[t]
\centering
\small
\begin{tabular}{lr}
\toprule
Property of the 634 correct Soft-EX / incorrect Strict-EX cases & Value \\
\midrule
Prediction has strictly more fields than gold & 634 (100\%) \\
\quad Extra fields per case (median / mean / max) & 1 / 10.1 / 1076 \\
Gold result projects a single field & 531 (83.8\%) \\
Prediction returns document-shaped output (has an id field) & 325 (51.3\%) \\
Differ by row count & 0 (0\%) \\
Differ by a disagreeing value & 0 (0\%) \\
\bottomrule
\end{tabular}
\caption{Field-level analysis of the 634 cases where Soft-EX and Strict-EX disagree (Claude-Opus-4-5, with knowledge). Every disagreement is the prediction returning the correct answer plus extra fields.}
\label{tab:soft-ex-stats}
\end{table*}

\paragraph{Examples.}
Table~\ref{tab:soft-ex-examples} shows three representative cases.
In each, Strict-EX fails only because the prediction returns extra fields, while Soft-EX correctly credits the answer.

\begin{table*}[t]
\centering
\small
\begin{tabular}{p{0.30\linewidth} p{0.28\linewidth} p{0.34\linewidth}}
\toprule
\textbf{Request} & \textbf{Ground-truth result} & \textbf{Prediction result (extra fields)} \\
\midrule
Where is the Malaysian Grand Prix held? Give the location coordinates. &
\texttt{[\{lat: 2.76083, lng: 101.738\}]} &
\texttt{[\{location: "Kuala Lumpur", lat: 2.76083, lng: 101.738\}]} \newline
\emph{one extra context field} \\
\addlinespace
For all drivers who finished race No.~872, who is the youngest? &
\texttt{[\{forename: "Sergio", surname: "P\'{e}rez"\}]} &
\texttt{[\{\_id: "21521", driverId: "815", forename: "Sergio", surname: "P\'{e}rez", dob: 1990-01-26\}]} \newline
\emph{whole driver document} \\
\addlinespace
For the only business in ``Yuma'' city, how many ``medium'' reviews did it get? &
\texttt{[\{count: 4\}]} &
\texttt{[\{\_id: "5969", medium\_review\_count: 4\}]} \newline
\emph{aggregate value plus id} \\
\bottomrule
\end{tabular}
\caption{Representative disagreement cases (Claude-Opus-4-5, with knowledge). Each prediction is scored incorrect by Strict-EX and correct by Soft-EX; in all three the prediction returns the requested answer together with additional document fields.}
\label{tab:soft-ex-examples}
\end{table*}

\section{Scaling Procedure for the Execution-Time Study}
\label{app:scaling}

The execution-time study (Fig.~\ref{fig:exec-scale}) measures how ground-truth MQL latency grows with database size under each conversion, across two benchmarks: three Spider databases (with all three conversions) and eleven BIRD databases (with Access-Pattern and One-to-One only). 
One-to-One queries are reproduced by following DocSpider's released conversion prompt. 
For FK-Embedding, although we reproduce TEND-v2's database migration, TEND-v2 does not release its SQL-to-MQL translation---neither the ground-truth MQL queries nor the conversion prompt---so FK-Embedding queries cannot be reproduced for the BIRD databases, and the timing study excludes it there.
To isolate the effect of \emph{schema design} from that of \emph{data volume}, we scale each source database while holding its relational schema, keys, and relationships fixed, then migrate the identical data under each conversion and time the same queries.

\paragraph{Synthetic proportional scale-up.}
We enlarge a database by a factor $k$ by \emph{appending} new synthetic rows rather than duplicating existing ones.
Every table receives $(k{-}1)\times$ its original row count, so all tables grow proportionally.
Each new row is built as follows: its primary key continues the integer sequence; each foreign key is assigned a uniformly random \emph{valid} parent key (parents are generated before their children, so every reference resolves); columns with a uniqueness constraint are generated from a monotonic counter; and all remaining values are sampled from the column's empirical distribution in the source data, so value ranges and byte sizes stay realistic.
Special cases are handled explicitly: one-to-one links (e.g.\ a fact row keyed to its parent), text-typed foreign keys that reference integer keys, and many-to-many junction tables.

\paragraph{Preserving undeclared relationships.}
Some source relationships are semantically real but not declared as foreign keys.
We preserve them so the scaled data remains faithful: for example, in \texttt{bike\_1}, \texttt{trip.start\_station\_id} and \texttt{end\_station\_id} reference \texttt{station.id} (new trips draw valid stations and keep the station name consistent with the id), and \texttt{weather(date, zip\_code)} is a composite key (new rows use fresh dates crossed with the existing zips, so the pair stays unique).

\paragraph{Integrity checks and target sizes.}
After scaling, each database passes SQLite's \texttt{PRAGMA foreign\_key\_check} and \texttt{PRAGMA integrity\_check} with no new violations introduced by generation.
For Spider we choose $k$ per database so that the migrated MongoDB database reaches roughly 50 and 200~MB, alongside its original size.
For BIRD we apply fixed nominal expansions of $5\times$ and $10\times$, alongside each database's original size; because BIRD databases start at very different sizes, these expansions span mean sizes of roughly 20, 80, and 160~MB across the eleven databases.
In all cases we verify per-document sizes stay within MongoDB's 16~MB limit before migrating.

\begin{figure*}[!ht]
    \centering
    \begin{subfigure}{\linewidth}
        \centering
        \includegraphics[width=\linewidth]{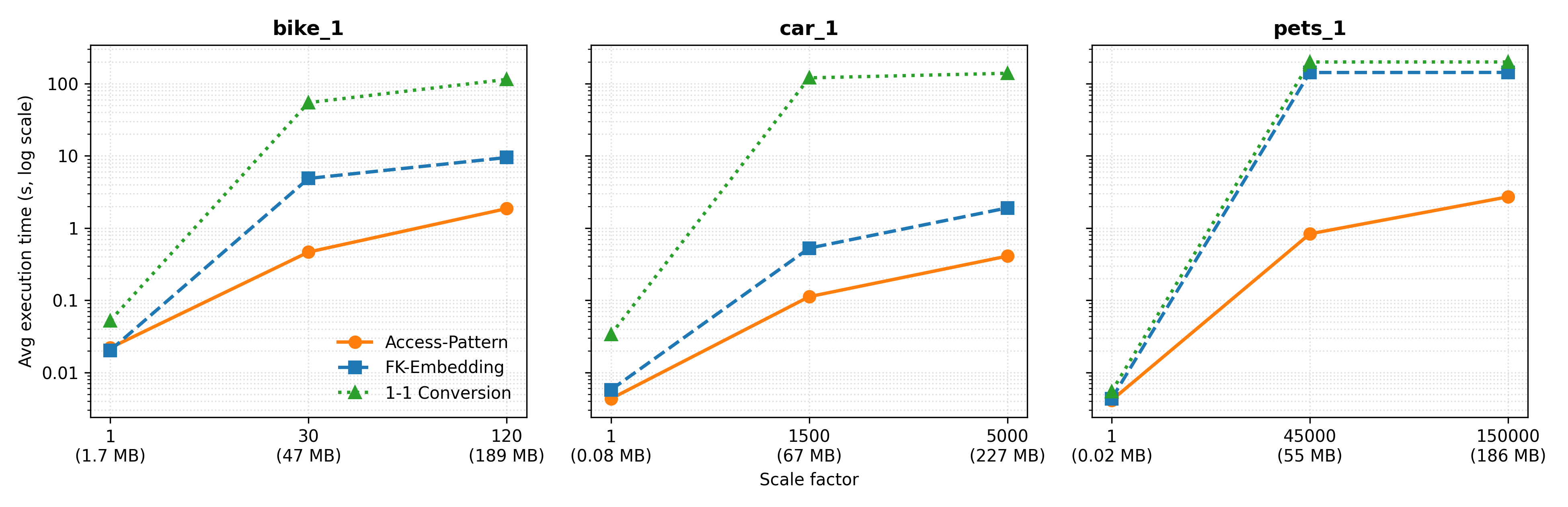}
        \caption{Spider databases.}
        \label{fig:exec-time-spider}
    \end{subfigure}

    \vspace{0.75em}

    \begin{subfigure}{\linewidth}
        \centering
        \includegraphics[width=\linewidth]{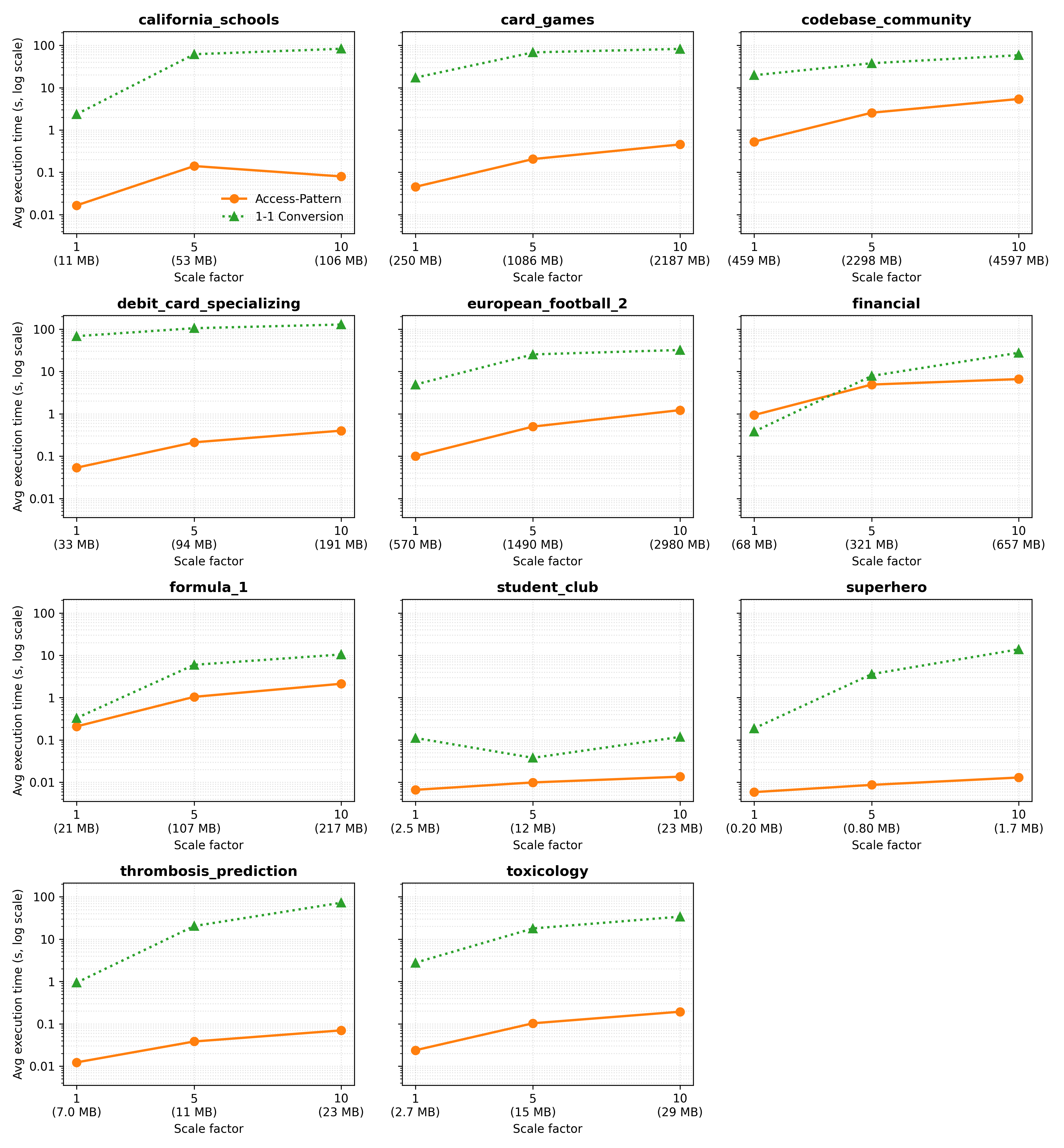}
        \caption{BIRD databases.}
        \label{fig:exec-time-bird}
    \end{subfigure}

    \caption{Average query execution time (log scale) versus scale factor for Access-Pattern, FK-Embedding, and 1-1 Conversion across the Spider and BIRD benchmarks.}
    \label{fig:exec-scale-appendix}
\end{figure*}

\paragraph{Migration and measurement.}
Each scaled SQLite database is migrated to MongoDB once per conversion under study: three times for Spider (one-to-one, foreign-key embedding, and access-pattern) and twice for BIRD (one-to-one and access-pattern).
We then run that database's ground-truth MQL queries against each migrated instance and record execution time per query, reporting the average over the query set at each size point (per-database results in Fig.~\ref{fig:exec-scale-appendix}).
The summary in Fig.~\ref{fig:exec-scale} aggregates these per-database averages using the geometric mean across databases, so that databases spanning several orders of magnitude in absolute latency contribute proportionally.
All measurements are taken on a MacBook Air (Apple M2, 24~GB RAM).

\input{error_analysis/acceptable_prediction_dev_1289}
\input{error_analysis/misunderstanding_database_content_dev_0871}
\input{error_analysis/misunderstanding_knowledge_evidence_dev_1052}
\input{error_analysis/wrong_schema_linking_dev_1419}
\input{error_analysis/wrong_calculation_dev_0452}


\end{document}

%% file: error_analysis/acl_visualization_macros.tex
\definecolor{acceptablebg}{HTML}{FFF2B2}
\definecolor{knowledgebg}{HTML}{FFD7C2}
\definecolor{contentbg}{HTML}{FFE3E3}
\definecolor{schemabg}{HTML}{DCEBFF}
\definecolor{calcbg}{HTML}{E6DDFF}
\definecolor{granularitybg}{HTML}{DDF7E8}
\definecolor{codegray}{HTML}{F7F7F7}

\newcommand{\categorybadgefull}[2]{%
  \noindent\colorbox{#1}{\parbox{\dimexpr\textwidth-2\fboxsep}{\textbf{#2}}}%
}
\newcommand{\noteboxfull}[2]{%
  \noindent\colorbox{#1}{\parbox{\dimexpr\textwidth-2\fboxsep}{#2}}%
}

\lstdefinestyle{mqlacl}{
  basicstyle=\ttfamily\scriptsize,
  breaklines=true,
  columns=fullflexible,
  keepspaces=true,
  showstringspaces=false,
  backgroundcolor=\color{codegray},
  frame=single,
  rulecolor=\color{black!18},
  xleftmargin=0.1em,
  xrightmargin=0.1em
}
\lstdefinestyle{resultacl}{
  basicstyle=\ttfamily\tiny,
  breaklines=true,
  columns=fullflexible,
  keepspaces=true,
  showstringspaces=false,
  backgroundcolor=\color{codegray},
  frame=single,
  rulecolor=\color{black!18}
}

%% file: error_analysis/acceptable_prediction_dev_1289.tex
\begin{table*}[p]
\centering
\small

\categorybadgefull{acceptablebg}{Acceptable Prediction (37\%) --- dev\_1289 / thrombosis\_prediction}
\vspace{0.45em}
\hrule
\vspace{0.55em}

\noindent\textbf{Question:}

For the patients who are diagnosed with SJS, how many of them have a normal level of total protein?

\medskip
\noindent\textbf{MongoDB Evidence:}

diagnosed with SJS refers to laboratories.patient.diagnosis = 'SJS'; normal level of total protein refers to laboratories.tp > 6.0 and laboratories.tp < 8.5;

\medskip
\noindent\begin{minipage}[t]{0.48\textwidth}
\textbf{Gold MQL:}
\begin{lstlisting}[style=mqlacl]
db.laboratories.aggregate(
[
  {
    "$match": {
      "patient.diagnosis": "SJS",
      "tp": {
        "$gt": 6.0,
        "$lt": 8.5
      }
    }
  },
  {
    "$count": "count"
  }
]
)
\end{lstlisting}
\end{minipage}
\hfill
\begin{minipage}[t]{0.48\textwidth}
\textbf{Prediction MQL:}
\begin{lstlisting}[style=mqlacl]
db.laboratories.aggregate(
[
  {
    "$match": {
      "patient.diagnosis": "SJS",
      "tp": {
        "$gt": 6.0,
        "$lt": 8.5
      }
    }
  },
  {
    "$group": {
      "_id": "$patientId"
    }
  },
  {
    "$count": "count"
  }
]
)
\end{lstlisting}
\end{minipage}

\vspace{0.55em}
\noteboxfull{acceptablebg}{\textbf{Key Difference:} The prediction counts distinct SJS patients with normal total protein, matching ``how many patients''; the gold counts lab rows.}

\medskip
\noindent\begin{minipage}[t]{0.48\textwidth}
\textbf{Gold Result:}
\begin{lstlisting}[style=resultacl]
[{"count": 780}]
\end{lstlisting}
\end{minipage}
\hfill
\begin{minipage}[t]{0.48\textwidth}
\textbf{Prediction Result:}
\begin{lstlisting}[style=resultacl]
[{"count": 37}]
\end{lstlisting}
\end{minipage}

\vspace{0.45em}
\hrule
\vspace{0.35em}
\caption{A representative acceptable prediction: the output fails exact matching but is semantically defensible.}
\label{tab:acceptable_prediction_dev_1289}
\end{table*}

%% file: error_analysis/misunderstanding_database_content_dev_0871.tex
\begin{table*}[p]
\centering
\small

\categorybadgefull{contentbg}{Misunderstanding Database Content (8\%) --- dev\_0871 / formula\_1}
\vspace{0.45em}
\hrule
\vspace{0.55em}

\noindent\textbf{Question:}

For the driver who had the Q2 time as 0:01:15 in race No. 347, where is he from?

\medskip
\noindent\textbf{MongoDB Evidence:}

race number refers to qualifying.raceId;

\medskip
\noindent\begin{minipage}[t]{0.48\textwidth}
\textbf{Gold MQL:}
\begin{lstlisting}[style=mqlacl]
db.qualifying.aggregate(
[
  {
    "$match": {
      "raceId": "347",
      "q2": {
        "$regularExpression": {
          "pattern": "^1:15",
          "options": ""
        }
      }
    }
  },
  {
    "$group": {
      "_id": "$driver.nationality"
    }
  },
  {
    "$project": {
      "_id": 0,
      "nationality": "$_id"
    }
  }
]
)
\end{lstlisting}
\end{minipage}
\hfill
\begin{minipage}[t]{0.48\textwidth}
\textbf{Prediction MQL:}
\begin{lstlisting}[style=mqlacl]
db.qualifying.aggregate(
[
  {
    "$match": {
      "raceId": "347",
      "q2": "0:01:15"
    }
  },
  {
    "$project": {
      "_id": 0,
      "nationality": "$driver.nationality"
    }
  }
]
)
\end{lstlisting}
\end{minipage}

\vspace{0.55em}
\noteboxfull{contentbg}{\textbf{Key Difference:} The prediction searches for the literal string `0:01:15`; the canonical query uses the stored Formula 1 time format prefix `1:15`.}

\medskip
\noindent\begin{minipage}[t]{0.48\textwidth}
\textbf{Gold Result:}
\begin{lstlisting}[style=resultacl]
[{"nationality": "Spanish"}, {"nationality": "Japanese"}, {"nationality": "German"}, {"nationality": "Russian"}, {"nationality": "Swiss"}]
\end{lstlisting}
\end{minipage}
\hfill
\begin{minipage}[t]{0.48\textwidth}
\textbf{Prediction Result:}
\begin{lstlisting}[style=resultacl]
[]
\end{lstlisting}
\end{minipage}

\vspace{0.45em}
\hrule
\vspace{0.35em}
\caption{A representative misunderstanding of database content: the prediction misses a dataset-specific value or stored representation.}
\label{tab:misunderstanding_database_content_dev_0871}
\end{table*}

%% file: error_analysis/misunderstanding_knowledge_evidence_dev_1052.tex
\begin{table*}[p]
\centering
\small

\categorybadgefull{knowledgebg}{Misunderstanding Knowledge Evidence (12\%) --- dev\_1052 / european\_football\_2}
\vspace{0.45em}
\hrule
\vspace{0.55em}

\noindent\textbf{Question:}

Among all the players whose weight is under 130, how many of them preferred foot in attacking is left?

\medskip
\noindent\textbf{MongoDB Evidence:}

players.weight < 130; preferred foot in attacking refers to player\_attributes.preferredFoot; player\_attributes.preferredFoot = 'left';

\medskip
\noindent\begin{minipage}[t]{0.48\textwidth}
\textbf{Gold MQL:}
\begin{lstlisting}[style=mqlacl]
db.players.aggregate(
[
  {
    "$match": {
      "weight": {
        "$lt": 130
      }
    }
  },
  {
    "$lookup": {
      "from": "player_attributes",
      "localField": "_id",
      "foreignField": "playerApiId",
      "as": "attributes"
    }
  },
  {
    "$match": {
      "attributes.preferredFoot": "left"
    }
  },
  {
    "$count": "count"
  }
]
)
\end{lstlisting}
\end{minipage}
\hfill
\begin{minipage}[t]{0.48\textwidth}
\textbf{Prediction MQL:}
\begin{lstlisting}[style=mqlacl]
db.players.aggregate(
[
  {
    "$match": {
      "weight": {
        "$lt": 130
      },
      "latestAttributes.preferredFoot": "left"
    }
  },
  {
    "$count": "count"
  }
]
)
\end{lstlisting}
\end{minipage}

\vspace{0.55em}
\noteboxfull{knowledgebg}{\textbf{Key Difference:} `MongoDB\_evidence` points to historical `player\_attributes.preferredFoot`; the prediction uses `players.latestAttributes.preferredFoot`, changing the candidate set.}

\medskip
\noindent\begin{minipage}[t]{0.48\textwidth}
\textbf{Gold Result:}
\begin{lstlisting}[style=resultacl]
[{"count": 9}]
\end{lstlisting}
\end{minipage}
\hfill
\begin{minipage}[t]{0.48\textwidth}
\textbf{Prediction Result:}
\begin{lstlisting}[style=resultacl]
[{"count": 7}]
\end{lstlisting}
\end{minipage}

\vspace{0.45em}
\hrule
\vspace{0.35em}
\caption{A representative misunderstanding of MongoDB evidence: the prediction does not transfer the provided path or mapping correctly.}
\label{tab:misunderstanding_knowledge_evidence_dev_1052}
\end{table*}

%% file: error_analysis/wrong_schema_linking_dev_1419.tex
\begin{table*}[p]
\centering
\small

\categorybadgefull{schemabg}{Wrong Schema Linking (6\%) --- dev\_1419 / student\_club}
\vspace{0.45em}
\hrule
\vspace{0.55em}

\noindent\textbf{Question:}

What is the category of event which was taken place in 2020-03-24T12:00:00?

\medskip
\noindent\textbf{MongoDB Evidence:}

taken place in 2020-03-24T12:00:00 refers to events.event\_date = '2020-03-24T12:00:00'

\medskip
\noindent\begin{minipage}[t]{0.48\textwidth}
\textbf{Gold MQL:}
\begin{lstlisting}[style=mqlacl]
db.events.aggregate(
[
  {
    "$match": {
      "event_date": "2020-03-24T12:00:00"
    }
  },
  {
    "$project": {
      "_id": 0,
      "category": "$budgets.category"
    }
  }
]
)
\end{lstlisting}
\end{minipage}
\hfill
\begin{minipage}[t]{0.48\textwidth}
\textbf{Prediction MQL:}
\begin{lstlisting}[style=mqlacl]
db.events.aggregate(
[
  {
    "$match": {
      "event_date": "2020-03-24T12:00:00"
    }
  },
  {
    "$project": {
      "_id": 0,
      "type": 1
    }
  }
]
)
\end{lstlisting}
\end{minipage}

\vspace{0.55em}
\noteboxfull{schemabg}{\textbf{Key Difference:} The prediction returns the event `type`; the canonical MQL asks for the event budget categories under `events.budgets.category`.}

\medskip
\noindent\begin{minipage}[t]{0.48\textwidth}
\textbf{Gold Result:}
\begin{lstlisting}[style=resultacl]
[{"category": ["Food", "Speaker Gifts", "Advertisement", "Parking"]}]
\end{lstlisting}
\end{minipage}
\hfill
\begin{minipage}[t]{0.48\textwidth}
\textbf{Prediction Result:}
\begin{lstlisting}[style=resultacl]
[{"type": "Guest Speaker"}]
\end{lstlisting}
\end{minipage}

\vspace{0.45em}
\hrule
\vspace{0.35em}
\caption{A representative wrong schema-linking case: the prediction uses a related but incorrect field or path.}
\label{tab:wrong_schema_linking_dev_1419}
\end{table*}

%% file: error_analysis/wrong_calculation_dev_0452.tex
\begin{table*}[p]
\centering
\small

\categorybadgefull{calcbg}{Wrong Calculation (33\%) --- dev\_0452 / card\_games}
\vspace{0.45em}
\hrule
\vspace{0.55em}

\noindent\textbf{Question:}

Please list the names of the cards that have a text box.

\medskip
\noindent\textbf{MongoDB Evidence:}

have a text box refers to cards.isTextless = false

\medskip
\noindent\begin{minipage}[t]{0.48\textwidth}
\textbf{Gold MQL:}
\begin{lstlisting}[style=mqlacl]
db.cards.aggregate(
[
  {
    "$match": {
      "isTextless": false
    }
  },
  {
    "$group": {
      "_id": "$name"
    }
  },
  {
    "$project": {
      "_id": 0,
      "name": "$_id"
    }
  },
  {
    "$sort": {
      "name": 1
    }
  }
]
)
\end{lstlisting}
\end{minipage}
\hfill
\begin{minipage}[t]{0.48\textwidth}
\textbf{Prediction MQL:}
\begin{lstlisting}[style=mqlacl]
db.cards.aggregate(
[
  {
    "$match": {
      "isTextless": false
    }
  },
  {
    "$project": {
      "_id": 0,
      "name": 1
    }
  }
]
)
\end{lstlisting}
\end{minipage}

\vspace{0.55em}
\noteboxfull{calcbg}{\textbf{Key Difference:} The prediction lists every non-textless printing name; the canonical MQL deduplicates to distinct card names.}

\medskip
\noindent\begin{minipage}[t]{0.48\textwidth}
\textbf{Gold Result:}
\begin{lstlisting}[style=resultacl]
[{"name": "\"Ach! Hans, Run!\""}, {"name": "\"Rumors of My Death . . .\""}, {"name": "1996 World Champion"}, {"name": "A Display of My Dark Power"}, {"name": "A Good Thing"}, {"name": "A Reckoning Approaches"}, {"name": "AWOL"}, {"name": "Abandon Hope"}, {"name": "Abandon Reason"}, {"name": "Abandoned Outpost"}, {"name": "Abandoned Sarcophagus"}, {"name": "Abattoir Ghoul"}, {"name": "Abbey Gargoyles"}, {"name": "Abbey Griffin"}, {"name": "Abbey Matron"}, {"name": "Abbot of Keral Keep"}, {"name": "Abduction"}, {"name": "Aberrant Researcher // Perfected Form"}, {"name": "Abeyance"}, {"name": "Abhorrent Overlord"},  ...
\end{lstlisting}
\end{minipage}
\hfill
\begin{minipage}[t]{0.48\textwidth}
\textbf{Prediction Result:}
\begin{lstlisting}[style=resultacl]
[{"name": "Ancestor's Chosen"}, {"name": "Angel of Mercy"}, {"name": "Aven Cloudchaser"}, {"name": "Ballista Squad"}, {"name": "Bandage"}, {"name": "Beacon of Immortality"}, {"name": "Benalish Knight"}, {"name": "Benalish Knight"}, {"name": "Cho-Manno, Revolutionary"}, {"name": "Condemn"}, {"name": "Contaminated Bond"}, {"name": "Drudge Skeletons"}, {"name": "Demystify"}, {"name": "Fear"}, {"name": "Field Marshal"}, {"name": "Hypnotic Specter"}, {"name": "Ghost Warden"}, {"name": "Glorious Anthem"}, {"name": "Hail of Arrows"}, {"name": "Heart of Light"}, {"name": "High Ground"}, {"name": "Honor Guard"}, {"name":  ...
\end{lstlisting}
\end{minipage}

\vspace{0.45em}
\hrule
\vspace{0.35em}
\caption{A representative wrong-calculation case: the prediction changes the aggregation level, distinctness, or formula.}
\label{tab:wrong_calculation_dev_0452}
\end{table*}